\documentclass[10pt,twocolumn,letterpaper]{article}

\usepackage[algorithms,pagenumbers]{wacv}

\usepackage{amsmath}
\usepackage{amssymb}
\usepackage{booktabs}
\usepackage{graphicx}
\usepackage{colortbl}
\usepackage{pifont}
\usepackage[accsupp]{axessibility}

\definecolor{wacvblue}{rgb}{0.21,0.49,0.74}
\usepackage[pagebackref,breaklinks,colorlinks,allcolors=wacvblue]{hyperref}

\def\wacvPaperID{1593}
\def\confName{WACV}
\def\confYear{2027}

\newcommand{\ours}{TAR}
\newcommand{\best}[1]{\textbf{#1}}
\definecolor{updeltacolor}{rgb}{0.0,0.45,0.0}
\definecolor{downdeltacolor}{rgb}{0.45,0.0,0.0}
\newcommand{\up}[1]{{\scriptsize\textcolor{updeltacolor}{(+#1)}}}
\newcommand{\dn}[1]{{\scriptsize\textcolor{downdeltacolor}{($-$#1)}}}

\title{Task-Aware Reasoning for Unified Multi-Task Visual Grounding}

\author{Sheng Cheng \qquad Wenhao Wu \qquad Bolin Lai \qquad Xiaofeng Gao \qquad Davide Modolo\\
Amazon AGI\\
{\tt\small \{cshengc, wenhaowu, bolinlai, gxiaofen, dmodolo\}@amazon.com}
}

\begin{document}
\maketitle
\begin{abstract}
Visual grounding spans a family of tasks, including spatial grounding, temporal grounding, spatio-temporal grounding, tracking, and segmentation.
Recent work has shown that reinforcement learning can train these grounding tasks jointly within a single multimodal large language model.
However, existing approaches normalize each task separately, placing related grounding tasks on different advantage scales.
We propose Task-Aware Reasoning (\ours{}), which combines unified advantage normalization with a generated task declaration.
We use a shared normalization denominator for all visual grounding tasks, placing their advantage magnitudes on a common reference scale.
Since each grounding task requires its own reasoning process and output schema, the model first generates a task declaration that conditions all subsequent reasoning and output.
Extensive experiments show that \ours{} achieves state-of-the-art performance among generalist models on most grounding benchmarks.
Across representative benchmarks, \ours{} delivers gains of +4.4 AO on GOT-10k, +5.2 R@0.5 on ActivityNet, +3.3 cIoU on RefCOCOg, and +4.8 J\&F on ReasonVOS over prior generalist models, while remaining competitive on general image and video QA.
\end{abstract}

\section{Introduction}

\begin{figure*}[t]
\centering
\includegraphics[width=0.8\textwidth]{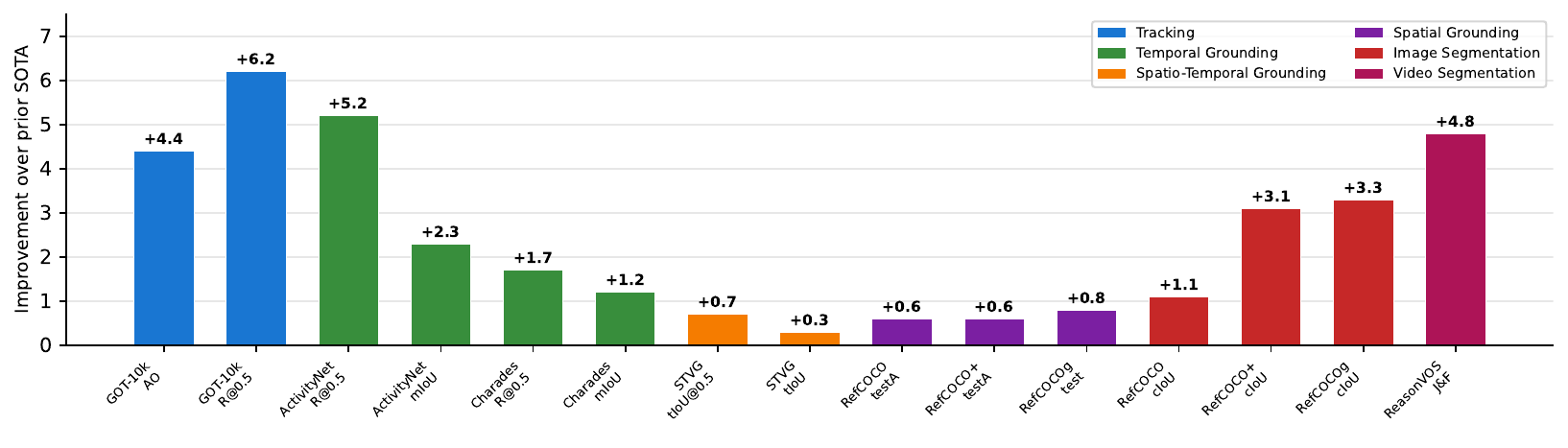}
\caption{Performance improvements on visual grounding tasks over OneThinker~\cite{feng2025onethinker}, which uses per-task normalization groups.}
\label{fig:results}
\end{figure*}

Visual grounding links natural-language descriptions to concrete visual evidence, encompassing spatial localization, temporal localization, spatio-temporal localization, object tracking, and referring segmentation across images and videos.
Multimodal large language models (MLLMs) provide a natural interface for this problem because they can interpret open-ended language instructions, reason over visual context, and generate structured localization outputs.
Recent MLLM methods have been developed for individual grounding tasks, including spatial grounding~\citep{vlmr1, perceptionr1}, temporal grounding~\citep{timer1, temporalrlt}, spatio-temporal grounding~\citep{groundinggpt, groundedvideollm}, segmentation~\citep{segr1}, and tracking~\citep{r1track}.
These methods demonstrate strong task-specific grounding ability, but they largely optimize each grounding task separately.

Since visual grounding tasks share underlying localization capabilities, recent work begins to train unified MLLMs across multiple image and video tasks to enable cross-task knowledge transfer and improved generalization.
For example, VideoChat-R1~\citep{videochatr1} co-trains on multiple spatio-temporal perception tasks, and OneThinker~\citep{feng2025onethinker} trains a single MLLM across ten image and video tasks.
A central challenge in such unified RL training is reward heterogeneity, since grounding tasks are scored by localization-quality rewards such as IoU while QA tasks use exact-match or reward-model scores.
OneThinker addresses this challenge with EMA-GRPO, which normalizes each task's advantages using its own EMA reward standard deviation.
However, within visual grounding, the rewards are comparable. Whether built from box overlap, interval overlap, or point distances, they all measure localization quality on a common bounded scale.
Under per-task EMA, the same difference from the prompt-group mean reward produces a larger advantage magnitude for a rollout from a task with a smaller reward standard deviation.
As a result, the relative weights of comparable grounding signals in the policy objective depend on task-level reward statistics rather than on the rewards themselves.

To address this, we propose Task-Aware Reasoning (\ours{}), a unified multi-task grounding framework with two complementary components.
First, during RL, all visual grounding tasks share one normalization denominator, so equal reward differences receive equal advantages across grounding tasks, while QA and captioning retain separate statistics.
Second, each grounding task requires its own reasoning process and output form.
We therefore have the model generate a task declaration at the start of its reasoning trace, identifying the task from the input so that all subsequent reasoning and the final answer are conditioned on it.
Together, the two components contribute on different levels, with shared normalization setting a common advantage scale during optimization and the declaration guiding task-specific generation.

We validate \ours{} through extensive experiments across six visual grounding task families and general multimodal QA (Figure~\ref{fig:results}).
\ours{} improves tracking by +4.4 AO (GOT-10k), temporal grounding by +5.2 R@0.5 (ActivityNet), image segmentation by +3.3 cIoU (RefCOCOg), and reasoning-intensive video segmentation by +4.8 J\&F (ReasonVOS).
On near-saturated spatial grounding benchmarks, the gains are smaller, while general video and image QA performance remains competitive.
Ablation studies further examine the contributions of unified normalization and task-aware reasoning.

Our main contributions are summarized as follows:
\begin{itemize}
\item We introduce unified advantage normalization, which uses a shared denominator across related visual grounding tasks and places their advantages on a common reference scale.
\item We propose task-aware reasoning for multi-task visual grounding, where the model generates an explicit task declaration before reasoning and receives a task-identification reward for its correctness. The declaration conditions subsequent reasoning and task-specific output.
\item Extensive experiments across six visual grounding task families and general multimodal QA demonstrate strong performance, with ablation studies examining both components.
\end{itemize}

\section{Related Work}
\label{sec:related}

\begin{figure*}[t]
\centering
\includegraphics[width=0.9\textwidth]{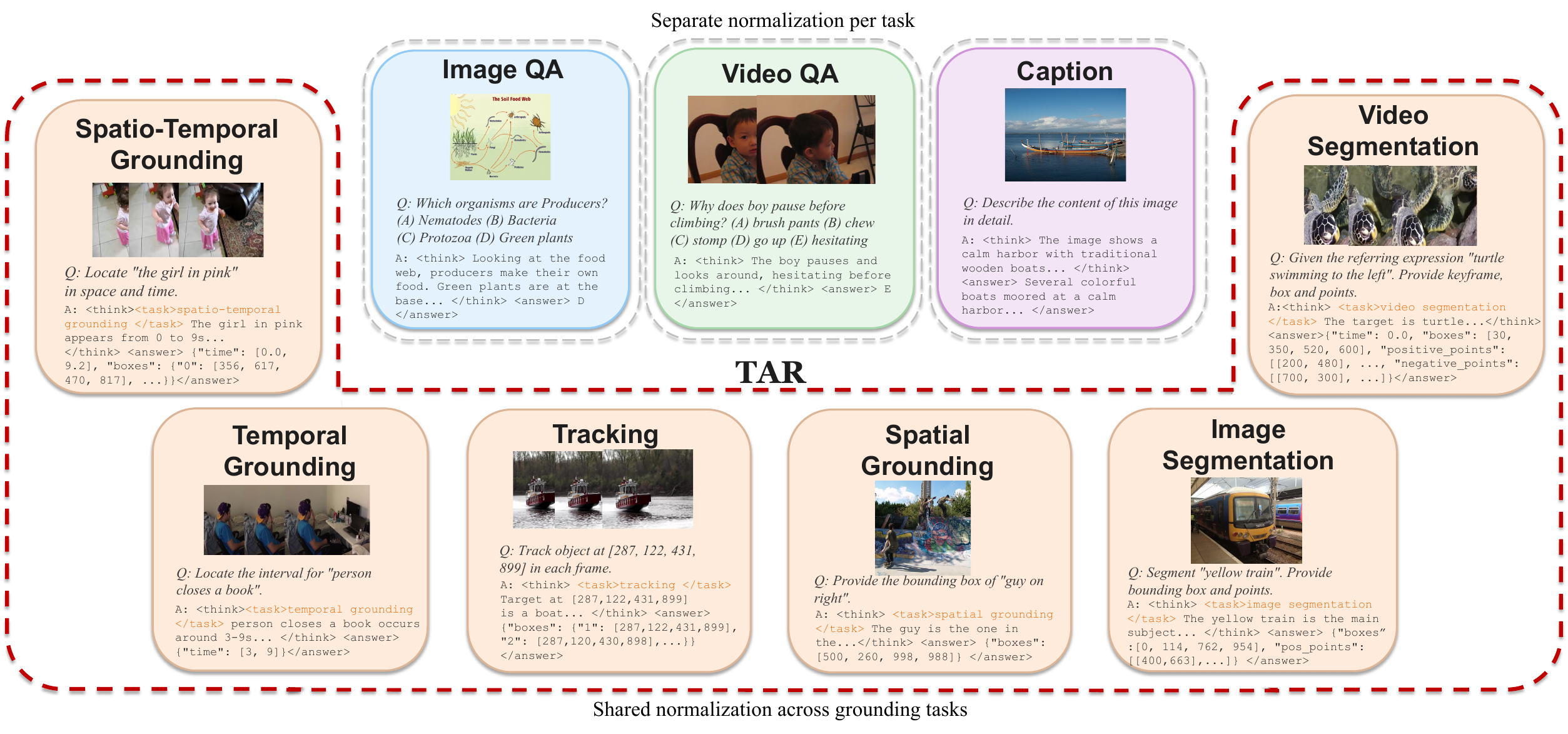}
\caption{Overview of our unified multi-task visual grounding framework. All grounding tasks (inside the red dashed box) share a single normalization group with task-aware reasoning, while QA and captioning tasks (inside gray dashed boxes) maintain independent normalization.}
\label{fig:teaser}
\end{figure*}

\paragraph{Visual grounding.}
Visual grounding spans a family of tasks that localize visual content described by natural language, differing only in the form of the target: a bounding box for spatial grounding~\citep{uninext, ferret, shikra}, a time interval for temporal grounding~\citep{timechat, vtimellm, timesuite, temporalrlt, timer1}, a box-within-an-interval for spatio-temporal grounding~\citep{groundinggpt, groundedvideollm}, a per-frame trajectory for single-object tracking~\citep{r1track}, and a pixel mask for referring segmentation in images~\citep{lisa, pixellm, segr1} and video~\citep{videolisa}. Despite this shared structure, these methods are almost always trained one task at a time. In contrast, we train a single model across all of them simultaneously and study how to optimize them jointly.

\paragraph{Multi-task visual reasoning with RL.}
Reinforcement learning has become a standard tool for improving visual reasoning in multimodal large language models (MLLMs), but most efforts target a single task or modality: image QA reasoning~\citep{sophiavl, visionr1, mmeureka, vlrethinker, vapothinker} or video understanding~\citep{videochatr1, videor1, longvilar1}. OneThinker~\citep{feng2025onethinker} is the first to train a single model across ten image and video tasks with RL, proposing EMA-GRPO to handle reward heterogeneity by normalizing GRPO advantages within each task separately. We adopt the same joint multi-task training, but differ in how advantages are normalized: rather than keeping a separate normalization group per task, we place related grounding tasks in a single shared GRPO normalization group, and add task-aware reasoning to further strengthen it. Conditioning a shared model on task identity is itself well studied: T5~\citep{raffel2020t5} prepends text prefixes and prompt tuning~\citep{lester2021power} learns task-specific soft tokens. Unlike these fixed input signals, our model \emph{generates} its own task declaration, which is therefore reinforced jointly with the task reward during RL rather than fixed at the input.

\paragraph{Reward normalization in multi-task RL.}
Normalizing rewards across tasks of differing scale is a core difficulty in multi-task RL~\citep{hessel2019multitask}. Standard GRPO~\citep{shao2024deepseekmath} divides advantages by the per-prompt-group standard deviation, introducing a question-level bias; Dr.GRPO~\citep{drgrpo} removes this division and leaves tasks on their raw reward scales; EMA-GRPO~\citep{feng2025onethinker} divides by per-task running statistics, placing each task on its own scale. We instead share one normalization scale across grounding tasks, whose rewards are comparable bounded localization measures, while other tasks keep separate statistics.

\section{Method}
\label{sec:method}

We address multi-task visual grounding, where a single model handles spatial grounding, temporal grounding, spatio-temporal grounding, tracking, and segmentation across images and videos. These tasks use related bounded localization rewards, but retain distinct reasoning processes and output schemas. We combine unified advantage normalization (\S\ref{sec:unified_ema}) with task-aware reasoning (\S\ref{sec:task_tag}), as illustrated in Figure~\ref{fig:teaser}.

\subsection{Task-Aware Reasoning}
\label{sec:task_tag}

Each grounding input implies a specific task and output schema. We make this task commitment explicit by generating a task declaration at the beginning of \texttt{<think>}, before free-form reasoning:
\begin{center}
\small
\fbox{\parbox{0.9\columnwidth}{
\texttt{<think><task>}$\tau$\texttt{</task>} \textit{reasoning...} \texttt{</think>}\\[2pt]
\texttt{<answer>} \textit{task-specific output} \texttt{</answer>}
}}
\end{center}
where $\tau \in \mathcal{T} = \{$\texttt{spatial grounding}, \texttt{temporal grounding}, \texttt{spatio-temporal grounding}, \texttt{tracking}, \texttt{image segmentation}, \texttt{video segmentation}$\}$ for grounding tasks. For non-grounding tasks (QA, captioning), the model omits the tag and proceeds directly with reasoning. Because the declaration is emitted before free-form reasoning, it conditions all subsequent reasoning and the final output on the declared task. Unlike input-side task prefixes~\citep{raffel2020t5}, the declaration is generated by the model at the start of its policy trajectory, making task routing part of the generated trajectory rather than an externally supplied condition.

During SFT, we inject the task tag into the beginning of the reasoning trace for all grounding samples and add a system instruction prompting the model to identify the task type before reasoning. At inference, the model autonomously decides whether and which tag to emit based on the input.

\subsection{Unified Advantage Normalization}
\label{sec:unified_ema}

Group relative policy optimization (GRPO) samples multiple rollouts per prompt and centers their rewards within each prompt group. EMA-GRPO~\citep{feng2025onethinker} normalizes a rollout $i$ from task $\tau$ using a task-wise reward standard deviation $\sigma_\tau(t)$ tracked by an exponential moving average:
\begin{equation}
A_i^{\mathrm{task}} = \frac{R_i - \bar{R}_g}{\sigma_\tau(t)},
\label{eq:task_adv}
\end{equation}
where $\bar{R}_g$ is the prompt-group mean reward. Each task is therefore standardized by its own reward dispersion. For grounding tasks with related bounded localization rewards, comparable centered reward differences can consequently receive different advantage magnitudes.

We instead pool all grounding tasks into a single normalization group and track one shared standard deviation $\sigma^{\mathrm{pool}}(t)$ via the same exponential moving average:
\begin{align}
m_1^{\mathrm{pool}}(t) &= \gamma\, m_1^{\mathrm{pool}}(t{-}1) + (1{-}\gamma)\, \bar{R}^{\mathrm{pool}}(t) \\
m_2^{\mathrm{pool}}(t) &= \gamma\, m_2^{\mathrm{pool}}(t{-}1) + (1{-}\gamma)\, \overline{R^2}^{\mathrm{pool}}(t)
\end{align}
where $\bar{R}^{\mathrm{pool}}(t)$ and $\overline{R^2}^{\mathrm{pool}}(t)$ are the mean reward and mean squared reward over all grounding-task rollouts in the batch at step $t$, and $\gamma{=}0.99$. The pooled standard deviation is $\sigma^{\mathrm{pool}}(t) = \sqrt{m_2^{\mathrm{pool}}(t) - (m_1^{\mathrm{pool}}(t))^2}$, and advantages are normalized as
\begin{equation}
A_i^{\mathrm{shared}} = \frac{R_i - \bar{R}_g}{\sigma^{\mathrm{pool}}(t)},
\label{eq:unified_adv}
\end{equation}
For the same rollout and prompt-group mean, the two advantages are related by
\begin{equation}
A_i^{\mathrm{shared}}
= A_i^{\mathrm{task}}\frac{\sigma_\tau(t)}{\sigma^{\mathrm{pool}}(t)}.
\label{eq:adv_relation}
\end{equation}
The shared denominator therefore removes task-specific rescaling and places all grounding advantages on a common reference scale. This changes their relative magnitudes across tasks, but does not equalize task gradients. The grounding tasks use related localization-quality measures bounded in $[0,1]$ for $R_{\mathrm{acc}}$ (Table~\ref{tab:rewards}), making a shared grounding statistic meaningful. QA and captioning retain separate statistics because they use different reward constructions.
\subsection{Reward Design}
\label{sec:reward}

Reinforcement learning requires a scalar reward for each response. Since the model is trained to emit a task declaration for grounding tasks (\S\ref{sec:task_tag}), our reward combines task accuracy, output format, and a task-identification term, composed as:
\begin{align}
R &= (1{-}\alpha)\cdot R_{\mathrm{base}} + \alpha \cdot R_{\mathrm{taskid}} \label{eq:total_reward} \\
R_{\mathrm{base}} &= (1{-}\lambda)\cdot R_{\mathrm{acc}} + \lambda \cdot R_{\mathrm{format}} + R_{\mathrm{struct}} \nonumber
\end{align}
where $R_{\mathrm{taskid}}$ rewards a correct task declaration (the \texttt{<task>} tag matching the ground-truth task), $R_{\mathrm{acc}}$ measures task-specific accuracy, $R_{\mathrm{format}}$ checks that the \texttt{<think>}/\texttt{<answer>} tags are well-formed, and $R_{\mathrm{struct}}$ is an additive bonus for a valid task-specific output schema inside \texttt{<answer>}. The schema bonus is gated by the format check: when format verification fails ($R_{\mathrm{format}}{=}0$), we set $R_{\mathrm{struct}}=0$, while the accuracy term $R_{\mathrm{acc}}$ is left unchanged. We set $\alpha = 0.1$ and $\lambda = 0.1$.

\paragraph{Task identification reward.}
SFT teaches the declaration format, while $R_{\mathrm{taskid}}$ maintains correct task commitment during stochastic RL. We define $R_{\mathrm{taskid}} \in \{0,1\}$ as a binary indicator of whether the model's declared task type $\hat{\tau}$, parsed from the generated \texttt{<task>} block, matches the ground-truth task label $\tau^*$. Unlike the task-accuracy reward, $R_{\mathrm{taskid}}$ directly evaluates declaration correctness. Once all rollouts in a prompt group declare the same correct task, this term becomes constant and cancels after group centering.

\begin{table}[t]
\centering
\resizebox{0.9\columnwidth}{!}{
\begin{tabular}{l l}
\toprule
\textbf{Task} & \textbf{$R_{\mathrm{acc}}$} \\
\midrule
Spatial grounding & $\mathrm{IoU}(\hat{b},\, b^*)$ \\
Temporal grounding & $\mathrm{tIoU}(\hat{t},\, t^*)$ \\
ST grounding & $\tfrac{1}{2}\mathrm{tIoU} + \tfrac{1}{2}\cdot\tfrac{1}{|F|}\sum_f \mathrm{IoU}(\hat{b}_f, b^*_f)$ \\
Tracking & $\tfrac{1}{|F|}\sum_f \mathrm{IoU}(\hat{b}_f, b^*_f)$ \\
Image segmentation & $\tfrac{1}{3}\mathrm{IoU}(\hat{b}, b^*) + \tfrac{1}{3}\mathcal{G}_\sigma(d_+) + \tfrac{1}{3}\mathcal{G}_\sigma(d_-)$ \\
Video segmentation & $\begin{aligned}[t]&\tfrac{1}{4}\mathrm{IoU}(\hat{b}, b^*) + \tfrac{1}{4}\mathcal{G}_1(|\hat{t}{-}t^*|)\\&+ \tfrac{1}{4}\mathcal{G}_\sigma(d_+) + \tfrac{1}{4}\mathcal{G}_\sigma(d_-)\end{aligned}$ \\
\midrule
Rule-based QA & Exact / symbolic match \\
Open-ended QA & Reward model score \\
\bottomrule
\end{tabular}}
\caption{Summary of task-specific accuracy rewards.}
\label{tab:rewards}
\end{table}
\begin{table*}[t]
\centering
\resizebox{0.7\textwidth}{!}{
\begin{tabular}{l c cccc cccc}
\toprule
& & \multicolumn{4}{c}{\textbf{Charades-STA}} & \multicolumn{4}{c}{\textbf{ActivityNet-Captions}} \\
\cmidrule(lr){3-6} \cmidrule(lr){7-10}
\textbf{Model} & \textbf{Frames} & R@0.3 & R@0.5 & R@0.7 & mIoU & R@0.3 & R@0.5 & R@0.7 & mIoU \\
\midrule
\multicolumn{10}{l}{\textit{Task-specific methods}} \\
VTimeLLM~\citep{vtimellm} & - & 55.3 & 34.3 & 14.7 & 34.6 & 44.8 & 29.5 & 14.2 & 31.4 \\
TimeSuite~\citep{timesuite} & - & 69.9 & 48.7 & 24.0 & - & - & - & - & - \\
Time-R1~\citep{timer1} & - & 78.1 & 60.8 & 35.3 & - & 58.6 & 39.0 & 21.4 & - \\
Temporal-RLT~\citep{temporalrlt} & - & 80.2 & 68.3 & 44.5 & 57.9 & 56.9 & 38.4 & 20.2 & 39.1 \\
VideoChat-R1~\citep{videochatr1} & - & 83.1 & 72.8 & 51.5 & 61.3 & 50.4 & 32.2 & 16.2 & 34.3 \\
\midrule
\multicolumn{10}{l}{\textit{Generalist models}} \\
Qwen3-VL-8B & 128 & 58.0 & 33.5 & 13.1 & 36.7 & 39.9 & 26.1 & 15.3 & 29.1 \\
OneThinker-8B~\citep{feng2025onethinker} & 128 & 83.5 & 68.3 & 45.3 & 59.9 & 65.0 & 43.6 & 25.7 & 45.9 \\
\rowcolor{gray!10}
\best{\ours{} (Ours)} & 128 & \best{84.0} & \best{70.0} & \best{47.5} & \best{61.1} & \best{68.7} & \best{48.8} & \best{28.1} & \best{48.2} \\
\bottomrule
\end{tabular}}
\caption{Performance on temporal grounding benchmarks. \textbf{Bold}: best among generalist models.}
\label{tab:temporal}
\end{table*}
\paragraph{Accuracy reward.}
$R_{\mathrm{acc}}$ measures how well the model's predicted output matches the ground-truth annotation. Since different grounding tasks produce different output types, each task requires a tailored reward function, summarized in Table~\ref{tab:rewards}.

\textbf{Spatial grounding.} The model predicts a bounding box $\hat{b} = [x_1, y_1, x_2, y_2]$ in normalized coordinates. The reward is the standard intersection-over-union: $R_{\mathrm{acc}} = \mathrm{IoU}(\hat{b}, b^*)$.

\textbf{Temporal grounding.} The model predicts a time interval $\hat{t} = [t_s, t_e]$. The reward is temporal IoU: $R_{\mathrm{acc}} = \mathrm{tIoU}(\hat{t}, t^*)$, measuring the overlap ratio between predicted and ground-truth intervals.

\textbf{Spatio-temporal grounding.} The model jointly predicts a temporal interval and per-frame bounding boxes. The reward averages a temporal and a spatial term: $R_{\mathrm{acc}} = \tfrac{1}{2}\mathrm{tIoU}(\hat{t}, t^*) + \tfrac{1}{2}\cdot\frac{1}{|F|}\sum_{f \in F} \mathrm{IoU}(\hat{b}_f, b^*_f)$, where $F$ is the set of annotated frames.

\textbf{Tracking.} The model predicts bounding boxes for each frame in a video sequence. The reward is the mean spatial IoU across all annotated frames: $R_{\mathrm{acc}} = \frac{1}{|F|}\sum_{f \in F} \mathrm{IoU}(\hat{b}_f, b^*_f)$.

\textbf{Image segmentation.} The model generates segmentation prompts: a bounding box $\hat{b}$, positive points $P_+$, and negative points $P_-$. These prompts are fed to a frozen SAM2~\citep{sam2} to produce the final masks, and our benchmark metric (cIoU) is computed on those masks. During training, however, decoding masks for every rollout is expensive, so we instead reward the quality of the prompts as a proxy for mask quality: $R_{\mathrm{acc}} = \tfrac{1}{3}\mathrm{IoU}(\hat{b}, b^*) + \tfrac{1}{3}\mathcal{G}_\sigma(d_+) + \tfrac{1}{3}\mathcal{G}_\sigma(d_-)$, where $\mathcal{G}_\sigma(\cdot)$ is a Gaussian kernel with bandwidth $\sigma{=}50$, and $d_+$ (resp.\ $d_-$) is the mean distance from the positive (resp.\ negative) points to the ground-truth mask region. Positive points inside the mask and negative points outside it receive higher reward.

\textbf{Video segmentation.} Extends image segmentation with a temporal keyframe component: $R_{\mathrm{acc}} = \tfrac{1}{4}\mathrm{IoU}(\hat{b}, b^*) + \tfrac{1}{4}\mathcal{G}_1(|\hat{t} - t^*|) + \tfrac{1}{4}\mathcal{G}_\sigma(d_+) + \tfrac{1}{4}\mathcal{G}_\sigma(d_-)$, where the Gaussian kernel on keyframe distance uses bandwidth $\sigma{=}1$ to sharply penalize temporal misalignment.

\textbf{Rule-based QA.} For multiple-choice, math, and numerical questions, the reward is binary exact match or symbolic equivalence.

\textbf{Open-ended QA and captioning.} An external reward model (POLAR-7B~\citep{polar}) provides a continuous similarity score between the model's response and the reference answer.

\paragraph{Format reward.}
$R_{\mathrm{format}} \in \{0,1\}$ verifies that the response adheres to the required generation structure: it must contain properly paired \texttt{<think>...</think>} and \texttt{<answer>...</answer>} tags. This prevents degenerate outputs such as emitting raw predictions without reasoning traces.

\paragraph{Structure reward.}
$R_{\mathrm{struct}} \in \{0, 0.5\}$ is an additive bonus that rewards valid task-specific output schema beyond surface-level format: for example, that tracking outputs contain properly-formed JSON with frame-indexed bounding boxes, or that grounding outputs contain parseable coordinates within the expected range. This provides partial credit for structurally correct responses even when accuracy is low, giving the model a smoother learning signal early in training.

\subsection{Experimental Setup}

\paragraph{Base model and training pipeline.}
We use Qwen3-VL-Instruct-8B~\citep{qwen3vl} as the base model with a two-stage pipeline: SFT cold start followed by RL. The SFT stage uses 340k samples from OneThinker SFT dataset with injected \texttt{<task>} tags (1 epoch, learning rate $1{\times}10^{-5}$, batch size 32).\footnote{We filter out data with inappropriate or non-permissive licenses.} The RL stage uses the OneThinker RL dataset~\citep{feng2025onethinker} with our unified EMA-GRPO (batch size 128, 8 rollouts per prompt, learning rate $2{\times}10^{-6}$, KL coefficient $\beta=0.01$, EMA decay $\gamma=0.99$, max response length 4096 tokens).

\paragraph{Evaluation protocol.}
All evaluations use greedy decoding with a maximum of 128 video frames. We evaluate across six grounding task families: spatial grounding (RefCOCO/+/g~\citep{refcoco}, accuracy at IoU 0.5, Acc@0.5), temporal grounding (Charades-STA~\citep{charades}, ActivityNet-Captions~\citep{activitynet}; recall at IoU threshold $k$, R@$k$, and mean IoU, mIoU), spatio-temporal grounding (STVG~\citep{stvg}; temporal and spatial IoU, tIoU and sIoU), tracking (GOT-10k~\citep{got10k}; average overlap, AO, and R@$k$), image segmentation (RefCOCO/+/g, cumulative IoU, cIoU), and video segmentation (MeViS~\citep{mevis} and ReasonVOS~\citep{videolisa}, region-and-boundary quality, J\&F). We additionally evaluate general video and image question answering to assess whether unified grounding training preserves broader multimodal reasoning.

\paragraph{Baselines.}
We compare against: (1)~OneThinker-8B~\citep{feng2025onethinker}, the current state of the art for unified multi-task visual grounding; (2)~Qwen3-VL-Instruct-8B, the base model; and (3)~task-specific specialist methods for each benchmark.\footnote{For all baselines, we use the values reported in their corresponding papers.}

\section{Main Results}
\label{sec:main_results}
We first evaluate our model across six visual grounding task families, including spatial grounding, temporal grounding, spatio-temporal grounding, tracking, image segmentation, and video segmentation. We then evaluate general video and image question answering to examine whether unified grounding training preserves broader multimodal reasoning ability. 

\begin{table}[t]
\centering
\resizebox{\columnwidth}{!}{
\begin{tabular}{l ccc ccc cc}
\toprule
& \multicolumn{3}{c}{\textbf{RefCOCO}} & \multicolumn{3}{c}{\textbf{RefCOCO+}} & \multicolumn{2}{c}{\textbf{RefCOCOg}} \\
\cmidrule(lr){2-4} \cmidrule(lr){5-7} \cmidrule(lr){8-9}
\textbf{Model} & testA & testB & val & testA & testB & val & test & val \\
\midrule
\multicolumn{9}{l}{\textit{Task-specific methods}} \\
Perception-R1~\citep{perceptionr1} & 91.4 & 84.5 & 89.1 & 86.8 & 74.3 & 81.7 & 85.4 & 85.7 \\
VLM-R1~\citep{vlmr1} & - & - & 90.5 & - & - & 84.3 & - & 87.1 \\
DeepEyes~\citep{deepeyes} & - & - & 89.8 & - & - & 83.6 & - & 86.7 \\
\midrule
\multicolumn{9}{l}{\textit{Generalist models}} \\
Qwen3-VL-8B & 92.2 & 85.3 & 89.9 & 89.6 & 77.8 & 84.5 & 86.7 & 86.8 \\
OneThinker-8B~\citep{feng2025onethinker} & 93.7 & 88.9 & 92.0 & 91.4 & 82.7 & 87.0 & 88.8 & 89.2 \\
\rowcolor{gray!10}
\best{\ours{} (Ours)} & \best{94.3} & \best{89.0} & \best{92.3} & \best{92.0} & \best{82.9} & \best{87.8} & \best{89.6} & \best{89.4} \\
\bottomrule
\end{tabular}}
\caption{Performance on spatial grounding (Acc@0.5).}
\label{tab:spatial}
\end{table}

\begin{table}[t]
\centering
\resizebox{\columnwidth}{!}{
\begin{tabular}{l c cccc}
\toprule
\textbf{Model} & \textbf{Frames} & tIoU@.5 & tIoU & sIoU@.5 & sIoU \\
\midrule
\multicolumn{6}{l}{\textit{Task-specific methods}} \\
GroundingGPT~\citep{groundinggpt} & - & 7.1 & 12.2 & 2.9 & 9.2 \\
VTimeLLM~\citep{vtimellm} & - & 7.1 & 15.5 & - & - \\
Grounded-VideoLLM~\citep{groundedvideollm} & - & 30.0 & 33.0 & - & - \\
\midrule
\multicolumn{6}{l}{\textit{Generalist models}} \\
Qwen3-VL-8B & 128 & 24.4 & 25.4 & 11.6 & 13.6 \\
OneThinker-8B~\citep{feng2025onethinker} & 128 & 34.9 & 39.5 & \best{40.3} & \best{36.7} \\
\rowcolor{gray!10}
\best{\ours{} (Ours)} & 128 & \best{35.6} & \best{39.8} & 38.1 & 36.4 \\
\bottomrule
\end{tabular}}
\caption{Performance on spatio-temporal grounding (STVG).}
\label{tab:stvg}
\end{table}

\begin{table}[t]
\centering
\resizebox{0.8\columnwidth}{!}{
\begin{tabular}{l cccc}
\toprule
\textbf{Model} & AO & R@0.3 & R@0.5 & R@0.7 \\
\midrule
\multicolumn{5}{l}{\textit{Task-specific methods}} \\
VideoChat-R1~\citep{videochatr1} & 42.5 & - & 30.6 & 3.9 \\
R1-Track~\citep{r1track} & 68.0 & - & 76.6 & - \\
\midrule
\multicolumn{5}{l}{\textit{Generalist models}} \\
Qwen3-VL-8B & 33.7 & 51.1 & 28.9 & 10.6 \\
OneThinker-8B~\citep{feng2025onethinker} & 73.0 & 93.9 & 84.4 & 68.8 \\
\rowcolor{gray!10}
\best{\ours{} (Ours)} & \best{77.4} & \best{96.1} & \best{90.6} & \best{75.6} \\
\bottomrule
\end{tabular}}
\caption{Performance on single-object tracking (GOT-10k).}
\label{tab:tracking}
\end{table}

\begin{table}[t]
\centering
\resizebox{0.8\columnwidth}{!}{
\begin{tabular}{l ccc}
\toprule
\textbf{Model} & RefCOCO & RefCOCO+ & RefCOCOg \\
\midrule
\multicolumn{4}{l}{\textit{Task-specific methods}} \\
LISA-7B~\citep{lisa} & 74.1 & 62.4 & 66.4 \\
PixelLM-7B~\citep{pixellm} & 73.0 & 66.3 & 69.3 \\
Seg-R1-7B~\citep{segr1} & 74.3 & 62.6 & 71.0 \\
\midrule
\multicolumn{4}{l}{\textit{Generalist models}} \\
Qwen3-VL-8B + SAM2 & 73.2 & 66.2 & 68.3 \\
OneThinker-8B~\citep{feng2025onethinker} & 75.8 & 67.1 & 70.8 \\
\rowcolor{gray!10}
\best{\ours{} (Ours)} & \best{76.9} & \best{70.2} & \best{74.1} \\
\bottomrule
\end{tabular}}
\caption{Performance on image segmentation (cIoU, val set).}
\label{tab:img_seg}
\end{table}

\begin{table}[t]
\centering
\resizebox{0.9\columnwidth}{!}{
\begin{tabular}{l ccc ccc}
\toprule
& \multicolumn{3}{c}{\textbf{MeViS}} & \multicolumn{3}{c}{\textbf{ReasonVOS}} \\
\cmidrule(lr){2-4} \cmidrule(lr){5-7}
\textbf{Model} & J & F & J\&F & J & F & J\&F \\
\midrule
\multicolumn{7}{l}{\textit{Task-specific methods}} \\
ReferFormer~\citep{referformer} & 29.8 & 32.2 & 31.0 & 30.2 & 35.6 & 32.9 \\
VISA-13B~\citep{visa} & - & - & 44.5 & - & - & - \\
VideoLISA-3.8B~\citep{videolisa} & 41.3 & 47.6 & 44.4 & 45.1 & 49.9 & 47.5 \\
\midrule
\multicolumn{7}{l}{\textit{Generalist models}} \\
OneThinker-8B~\citep{feng2025onethinker} & \best{48.8} & \best{56.7} & \best{52.7} & 51.1 & 58.7 & 54.9 \\
\rowcolor{gray!10}
\best{\ours{} (Ours)} & 48.1 & 56.0 & 52.0 & \best{55.8} & \best{63.7} & \best{59.7} \\
\bottomrule
\end{tabular}}
\caption{Performance on video segmentation benchmarks.}
\label{tab:vid_seg}
\end{table}

\begin{table*}[t]
\centering
\resizebox{0.65\textwidth}{!}{
\begin{tabular}{l c cccc}
\toprule
\textbf{Model} & \textbf{Frames} & \textbf{VideoMMMU} & \textbf{MMVU} & \textbf{VideoMME} & \textbf{LongVideo-Reason} \\
\midrule
\multicolumn{6}{l}{\textit{Proprietary models}} \\
GPT-4o~\citep{gpt4o} & - & 61.2 & 75.4 & 71.9 & - \\
Gemini 2.5 Pro~\citep{gemini25} & - & 83.6 & - & 84.3 & - \\
Seed1.5-VL~\citep{seed15vl} & - & 81.4 & - & 77.9 & - \\
\midrule
\multicolumn{6}{l}{\textit{Open-source task-specific}} \\
VideoChat-R1-7B~\citep{videochatr1} & - & 46.4 & - & 60.0 & 67.2 \\
LongVILA-R1-7B~\citep{longvilar1} & - & 51.0 & - & 65.1 & 72.0 \\
Video-R1-7B~\citep{videor1} & - & 52.4 & 64.2 & 61.4 & 68.1 \\
\midrule
\multicolumn{6}{l}{\textit{Generalist models}} \\
Qwen3-VL-8B & 128 & 63.3 & 65.6 & 64.0 & 71.5 \\
OneThinker-8B~\citep{feng2025onethinker} & 128 & 66.2 & \best{70.5} & 66.5 & 79.2 \\
\rowcolor{gray!10}
\best{\ours{} (Ours)} & 128 & \best{68.4} & 70.4 & \best{67.1} & \best{80.0} \\
\bottomrule
\end{tabular}}
\caption{Performance on video question answering benchmarks.}
\label{tab:videoqa}
\end{table*}

\subsection{Spatial Grounding}
As shown in Table~\ref{tab:spatial}, \ours{} achieves the best results across all eight evaluation splits of the RefCOCO family. Specifically, it reaches 94.3 on RefCOCO testA, 92.0 on RefCOCO+ testA, and 89.6 on RefCOCOg test. The improvements over OneThinker range from +0.1 to +0.8, so \ours{} further improves this near-saturated benchmark on every split. Compared to task-specific spatial grounding methods (Perception-R1, VLM-R1, DeepEyes), \ours{} outperforms all of them by substantial margins despite being trained for multiple tasks simultaneously.

\subsection{Temporal Grounding}

As shown in Table~\ref{tab:temporal}, \ours{} is the strongest generalist model on both benchmarks. On Charades-STA, it achieves 70.0 R@0.5 and 61.1 mIoU, improving over OneThinker by 1.7 points at R@0.5. On ActivityNet-Captions, which has longer videos and more diverse temporal patterns, the improvement is larger: \ours{} gains 5.2 points at R@0.5 and 2.3 points in mIoU over OneThinker, and exceeds the best task-specific method, Time-R1, by 9.8 points at R@0.5. It also outperforms VideoChat-R1 on ActivityNet by 16.6 points at R@0.5.

\subsection{Spatio-Temporal Grounding}

Table~\ref{tab:stvg} shows results on STVG. \ours{} is slightly ahead of OneThinker on the temporal metrics (tIoU 39.8 vs.\ 39.5, tIoU@0.5 35.6 vs.\ 34.9) and comparable on the spatial metrics (sIoU 36.4 vs.\ 36.7). Overall performance on this task is on par with OneThinker.

\subsection{Tracking}

As shown in Table~\ref{tab:tracking}, \ours{} achieves 77.4 AO, 96.1 R@0.3, 90.6 R@0.5, and 75.6 R@0.7, improving over OneThinker by 4.4 AO. The improvement increases as the overlap threshold becomes stricter, from 2.2 points at R@0.3 to 6.2 points at R@0.5 and 6.8 points at R@0.7, showing that \ours{} localizes the target more precisely in each frame. The gains at R@0.5 and R@0.7 are the largest single-metric improvements over OneThinker across all benchmarks. \ours{} also surpasses R1-Track, a dedicated tracking model, by 9.4 AO and 14.0 points at R@0.5.

\subsection{Image Segmentation}

As shown in Table~\ref{tab:img_seg}, \ours{} achieves 76.9, 70.2, and 74.1 cIoU on RefCOCO, RefCOCO+, and RefCOCOg, improving over OneThinker by 1.1, 3.1, and 3.3 points, respectively. It also outperforms all task-specific segmentation methods, including LISA-7B, PixelLM-7B, and Seg-R1-7B. The gains are larger on the two harder benchmarks: RefCOCO+ disallows location words, so its expressions rely on appearance, and RefCOCOg uses longer and more complex expressions.

\subsection{Video Segmentation}

Table~\ref{tab:vid_seg} shows video segmentation results. On MeViS, \ours{} achieves 52.0 J\&F, comparable to OneThinker at 52.7. On ReasonVOS, it reaches 59.7 J\&F, a 4.8-point improvement over OneThinker, and surpasses all task-specific methods, including VideoLISA-3.8B at 47.5. The two benchmarks test different abilities: MeViS describes targets mainly by their motion, whereas ReasonVOS requires linguistic reasoning to identify the referent, where \ours{} shows one of its largest gains.

\subsection{Video Question Answering}

Table~\ref{tab:videoqa} shows video QA results across four benchmarks. \ours{} achieves 68.4\% on VideoMMMU~\citep{videommmu}, 70.4\% on MMVU~\citep{mmvu}, 67.1\% on VideoMME~\citep{videomme}, and 80.0\% on LongVideo-Reason~\citep{longvilar1}, outperforming OneThinker on three of four benchmarks while matching on MMVU. These results surpass all open-source task-specific video reasoning models by large margins (e.g., +16.0 over Video-R1-7B on VideoMMMU). This indicates that unified grounding training preserves general video understanding, and the gains on three of four benchmarks suggest it may even reinforce it through shared visual reasoning.

\subsection{Image Question Answering}

\begin{table}[t]
\centering
\resizebox{0.8\columnwidth}{!}{
\begin{tabular}{l ccc}
\toprule
\textbf{Model} & \textbf{MMBench} & \textbf{MMStar} & \textbf{AI2D} \\
\midrule
\multicolumn{4}{l}{\textit{Proprietary models}} \\
GPT-4o~\citep{gpt4o} & 84.3 & 65.1 & 84.9 \\
Gemini 2.5 Pro~\citep{gemini25} & 90.1 & 77.5 & 88.4 \\
\midrule
\multicolumn{4}{l}{\textit{Open-source task-specific}} \\
SophiaVL-R1-7B~\citep{sophiavl} & 85.4 & 66.7 & - \\
VAPO-Thinker-7B~\citep{vapothinker} & - & 63.0 & - \\
\midrule
\multicolumn{4}{l}{\textit{Generalist models}} \\
Qwen3-VL-8B & 85.1 & 68.5 & 82.3 \\
OneThinker-8B~\citep{feng2025onethinker} & \best{86.6} & 70.6 & \best{85.2} \\
\rowcolor{gray!10}
\best{\ours{} (Ours)} & 86.3 & \best{71.1} & \best{85.2} \\
\bottomrule
\end{tabular}}
\caption{Performance on image question answering benchmarks.}
\label{tab:imageqa}
\end{table}

Table~\ref{tab:imageqa} shows image QA results. \ours{} achieves 86.3 on MMBench~\citep{mmbench}, 71.1 on MMStar~\citep{mmstar}, and 85.2 on AI2D~\citep{ai2d}, comparable to or exceeding OneThinker. This confirms that unified grounding training preserves image QA capabilities.

\section{Ablation Study}
\label{sec:ablation}

We ablate the task declaration and normalization design, compare normalization groupings and alternative optimization strategies, and examine the contribution of each training stage.

\begin{table}[t]
\centering
\resizebox{0.85\columnwidth}{!}{
\begin{tabular}{l ccc}
\toprule
\textbf{Benchmark} & \textbf{Qwen3-VL-8B} & \textbf{+ SFT} & \textbf{+ RL} \\
\midrule
\multicolumn{4}{l}{\textit{Spatial Grounding (Acc@0.5)}} \\
RefCOCO testA & 92.2 & 92.4~\up{0.2} & 94.3~\up{1.9} \\
RefCOCO+ testA & 89.6 & 88.2~\dn{1.4} & 92.0~\up{3.8} \\
RefCOCOg test & 86.7 & 88.7~\up{2.0} & 89.6~\up{0.9} \\
\midrule
\multicolumn{4}{l}{\textit{Temporal Grounding}} \\
Charades R@0.5 & 33.5 & 66.8~\up{33.3} & 70.0~\up{3.2} \\
\midrule
\multicolumn{4}{l}{\textit{Tracking}} \\
GOT-10k R@0.5 & 28.9 & 48.9~\up{20.0} & 90.6~\up{41.7} \\
\midrule
\multicolumn{4}{l}{\textit{Spatio-Temporal Grounding}} \\
STVG tIoU@0.5 & 24.4 & 23.2~\dn{1.2} & 35.6~\up{12.4} \\
\midrule
\multicolumn{4}{l}{\textit{Image Segmentation (cIoU)}} \\
RefCOCO & 73.2 & 75.5~\up{2.3} & 76.9~\up{1.4} \\
RefCOCO+ & 66.2 & 66.2~\up{0.0} & 70.2~\up{4.0} \\
RefCOCOg & 68.3 & 70.8~\up{2.5} & 74.1~\up{3.3} \\
\midrule
\multicolumn{4}{l}{\textit{Video Segmentation (J\&F)}} \\
MeViS & 22.9 & 48.3~\up{25.4} & 52.0~\up{3.7} \\
ReasonVOS & 19.7 & 55.9~\up{36.2} & 59.7~\up{3.8} \\
\midrule
\multicolumn{4}{l}{\textit{Video QA}} \\
VideoMMMU & 63.3 & 66.0~\up{2.7} & 68.4~\up{2.4} \\
\bottomrule
\end{tabular}}
\caption{Contribution of each training stage: base model (Qwen3-VL-8B) $\to$ SFT with task-aware data $\to$ RL with unified normalization and task-ID reward. Small colored numbers are the change of each stage over the previous one: \textbf{+ SFT} relative to Qwen3-VL-8B, and \textbf{+ RL} relative to \textbf{+ SFT}.}
\label{tab:progression}
\end{table}

\subsection{Effect of Unified Normalization and Task Declaration}

\begin{table*}[t]
\centering
\resizebox{0.75\textwidth}{!}{
\begin{tabular}{l cc c c c c c c}
\toprule
& \multicolumn{2}{c}{\textbf{Config}} & \textbf{Tracking} & \textbf{Temporal} & \textbf{STVG} & \textbf{Spatial} & \textbf{Seg (img)} & \textbf{Seg (vid)} \\
\cmidrule(lr){2-3} \cmidrule(lr){4-4} \cmidrule(lr){5-5} \cmidrule(lr){6-6} \cmidrule(lr){7-7} \cmidrule(lr){8-8} \cmidrule(lr){9-9}
\textbf{Method} & \textbf{Norm.} & \textbf{Tag} & AO & AN-R@0.5 & tIoU@0.5 & RC+-tA & RCg-cIoU & RVOS-J\&F \\
\midrule
OneThinker            & Per-task & \ding{55} & 73.0 & 43.6 & 34.9 & 91.4 & 70.8 & 54.9 \\
\ours{} {\footnotesize(w/o declaration)} & Shared & \ding{55} & 75.8 & 45.5 & 35.1 & 91.5 & 71.6 & 57.9 \\
\ours{} {\footnotesize(per-task)}     & Per-task & \ding{51} & 71.8 & 45.9 & \best{35.8} & 90.8 & 71.5 & 57.4 \\
\midrule
\best{\ours{}}        & Shared   & \ding{51} & \best{77.4} & \best{48.8} & 35.6 & \best{92.0} & \best{74.1} & \best{59.7} \\
\bottomrule
\end{tabular}}
\caption{Ablation on unified normalization and the task declaration.
\textbf{Norm.} denotes per-task or shared advantage normalization, and \textbf{Tag} indicates whether the model generates a task declaration.
Each column reports one representative benchmark: GOT-10k AO (tracking), ActivityNet R@0.5 (temporal), STVG tIoU@0.5, RefCOCO+ testA Acc@0.5 (spatial), RefCOCOg cIoU (image seg.), and ReasonVOS J\&F (video seg.).}

\label{tab:ablation_ema}
\end{table*}

\begin{table*}[t]
\centering
\resizebox{\textwidth}{!}{
\begin{tabular}{l l l c c c c c c}
\toprule
\textbf{Configuration} & \textbf{Normalization / optimization} & \textbf{Task declaration} & \textbf{Tracking} & \textbf{Temporal} & \textbf{STVG} & \textbf{Spatial} & \textbf{Seg (img)} & \textbf{Seg (vid)} \\
\cmidrule(lr){4-4} \cmidrule(lr){5-5} \cmidrule(lr){6-6} \cmidrule(lr){7-7} \cmidrule(lr){8-8} \cmidrule(lr){9-9}
& & & AO & AN-R@0.5 & tIoU@0.5 & RC+-tA & RCg-cIoU & RVOS-J\&F \\
\midrule
\multicolumn{9}{l}{\textit{Normalization and grouping variants}} \\
Dr.GRPO~\citep{drgrpo} & No reward std. & Generated + $R_{\mathrm{taskid}}$ & 77.2 & 47.8 & 32.9 & 91.0 & 72.2 & 59.0 \\
Coordinate / segmentation split & Two grounding EMAs & Generated + $R_{\mathrm{taskid}}$ & 73.2 & 46.5 & 32.7 & 91.3 & 71.6 & 59.1 \\
All RL tasks & One global EMA & Generated + $R_{\mathrm{taskid}}$ & 76.0 & 46.8 & 30.8 & 91.8 & 72.3 & 59.3 \\
\midrule
\multicolumn{9}{l}{\textit{Task-declaration variants}} \\
Generated without reward & One grounding EMA & Generated, $\alpha{=}0$ & 75.0 & 47.0 & 34.1 & 91.4 & 72.0 & 58.9 \\
Oracle input-side & One grounding EMA & Input GT & 76.8 & 48.4 & 34.9 & \best{92.1} & 71.3 & 58.8 \\
\midrule
\multicolumn{9}{l}{\textit{Multi-task optimization baseline}} \\
MT-GRPO~\citep{mtgrpo} & IWU + RP sampler & Generated + $R_{\mathrm{taskid}}$ & 76.2 & 48.0 & 30.7 & 91.3 & 71.9 & 57.9 \\
\midrule
\best{\ours{}} & One grounding EMA & Generated + $R_{\mathrm{taskid}}$ & \best{77.4} & \best{48.8} & \best{35.6} & 92.0 & \best{74.1} & \best{59.7} \\
\bottomrule
\end{tabular}}
\caption{Additional normalization, grouping, optimization, and task-declaration ablations. All configurations use the same base model, training data, and rollout budget. We report the same representative benchmarks as Table~\ref{tab:ablation_ema}.}
\label{tab:grouping_optimization}
\end{table*}

Table~\ref{tab:ablation_ema} isolates the two main components. The no-declaration variant tests shared normalization without explicit task routing, while the per-task variant tests the generated declaration under the original normalization scheme. Shared normalization provides the stronger overall result in both comparisons. The declaration further improves the shared-normalization model, while its mixed effect under per-task normalization shows that the two components are best understood as a coupled design.

\subsection{Additional Ablations}

\noindent\textbf{Normalization and grouping.}
Under the same base model, training data, and rollout budget, we retain the generated declaration and $R_{\mathrm{taskid}}$ while varying how advantages are normalized. Dr.GRPO removes reward-standard-deviation normalization. The coordinate/segmentation split assigns spatial, temporal, spatio-temporal grounding, and tracking to one EMA, and image and video segmentation to another. The global variant shares one EMA across all RL tasks, including QA. Each alternative degrades on all six representative metrics relative to our grounding EMA. These results validate a shared statistic for related grounding tasks rather than no normalization, finer fragmentation, or indiscriminate pooling.

\noindent\textbf{Task declaration.}
We fix one grounding EMA and vary how task identity is provided. Generated $\alpha{=}0$ retains the self-generated declaration but removes $R_{\mathrm{taskid}}$, while the oracle variant supplies the ground-truth task identity in the input without generation or $R_{\mathrm{taskid}}$. Without the reward, tag accuracy falls from 100\% after SFT to 98.5\% after RL and all six metrics decrease. The oracle improves over the unrewarded declaration on four metrics, but the complete model exceeds the oracle on five without requiring a task label at inference. This supports a generated and rewarded declaration as task routing rather than merely an externally supplied task identifier.

\noindent\textbf{Multi-task optimization.}
We compare with MT-GRPO under the same data and rollout budget to test whether improvement-aware weight updates and ratio-preserving sampling provide a stronger alternative to normalization grouping. MT-GRPO is lower than \ours{} on all six representative metrics, indicating that its adaptive task weighting does not recover the overall performance profile of the shared grounding scale.

\subsection{Contribution of Each Training Stage}

Table~\ref{tab:progression} shows the contribution of each training stage. The SFT stage serves mainly as a cold start that teaches the model the required output format and task declaration. After SFT, the model emits the correct \texttt{<task>} tag on 100\% of 100 randomly sampled evaluation cases per task. RL with unified normalization then delivers the main performance gains, most dramatically on tracking (48.9$\to$90.6 R@0.5) and STVG (23.2$\to$35.6 tIoU@0.5), while task-declaration accuracy remains at 100\% at the end of RL training.

\section{Conclusion}

We present \ours{}, a unified multi-task visual grounding framework with two complementary training mechanisms. Because related grounding tasks use comparable bounded localization rewards, unified advantage normalization places them on a common reference scale instead of rescaling each task separately. A generated and rewarded task declaration conditions subsequent reasoning and output for task-specific routing. Across six grounding task families, \ours{} improves most datasets and metrics, with substantial gains in tracking, temporal grounding, and image and video segmentation, while general QA performance remains broadly stable. Ablations support the shared grounding group and show that the declaration is most effective within this design. Overall, \ours{} provides a stronger grounding profile while remaining competitive on the few metrics without gains.
{\small
\bibliographystyle{ieeenat_fullname}
\bibliography{main}
}
\clearpage
\appendix
\onecolumn
\begin{center}
{\Large\bfseries Appendix}
\end{center}

\section{Can Unified Rewards Further Boost Performance?}

A natural follow-up question is whether further unifying the \emph{reward formulas} themselves can compound the gains from unified normalization. We design a single reward function, TubeIoU, that measures the overlap between predicted and ground-truth spatiotemporal tubes:
\begin{equation}
\mathrm{TubeIoU}(\hat{T}, T^*) = \frac{\sum_{t=1}^{N} |\hat{M}_t \cap M^*_t|}{\sum_{t=1}^{N} |\hat{M}_t \cup M^*_t|},
\end{equation}
where $\hat{M}_t$ and $M^*_t$ are the predicted and ground-truth regions at frame $t$, and $N$ is the total number of frames. A region is $\emptyset$ at frames outside the tube's temporal extent. This formulation computes IoU over the entire spatiotemporal volume in a single expression, subsuming spatial grounding ($N{=}1$), temporal grounding (full-frame regions where only interval overlap matters), and tracking (per-frame boxes over the full video). For segmentation tasks, we use SAM2 to materialize the model's emitted prompts into masks before computing IoU.
\begin{table}[h!]
\centering
\resizebox{0.75\textwidth}{!}{
\begin{tabular}{l c c c c c c}
\toprule
& \textbf{Tracking} & \textbf{Temporal} & \textbf{STVG} & \textbf{Spatial} & \textbf{Seg (img)} & \textbf{Seg (vid)} \\
\cmidrule(lr){2-2} \cmidrule(lr){3-3} \cmidrule(lr){4-4} \cmidrule(lr){5-5} \cmidrule(lr){6-6} \cmidrule(lr){7-7}
\textbf{Method} & AO & AN-R@0.5 & tIoU@0.5 & RC+-tA & RCg-cIoU & RVOS-J\&F \\
\midrule
\best{\ours{}} (heterogeneous) & \best{77.4} & \best{48.8} & 35.6 & \best{92.0} & \best{74.1} & \best{59.7} \\
w/ TubeIoU + SAM2 & 76.4 & 47.9 & \best{35.8} & 91.5 & 72.2 & 57.4 \\
\bottomrule
\end{tabular}}
\caption{Ablation on unified reward (TubeIoU + SAM2 vs.\ heterogeneous per-task rewards). TubeIoU is lower on all task families except STVG, indicating no benefit from reward-level unification. We report the same representative benchmarks as Table~\ref{tab:ablation_ema}.}
\label{tab:ablation_reward}
\end{table}

As shown in Table~\ref{tab:ablation_reward}, the unified TubeIoU reward does not improve over the heterogeneous per-task rewards, and is lower on every task family except STVG. TubeIoU rewards getting the temporal extent and the per-frame spatial overlap right jointly, which is most closely aligned with STVG's joint temporal-spatial objective. The other tasks do not jointly predict a temporal extent and spatial boxes, so coupling both axes into a single reward is less informative.
These results show that shared normalization does not require a unified reward formulation.

\section{Cross-Task Transfer under Task Removal}

We remove temporal-grounding examples from the RL mixture while retaining the same SFT
checkpoint, model, normalization, optimizer, compute, and remaining training data. This setting
tests whether training on the other grounding tasks can improve temporal grounding without direct
temporal-grounding examples during RL.
\begin{table}[h!]
\centering
\resizebox{0.45\textwidth}{!}{
\begin{tabular}{lc}
\toprule
\textbf{Training stage} & \textbf{ActivityNet R@0.5} \\
\midrule
SFT checkpoint & 26.7 \\
RL without temporal-grounding data & 38.3 \\
Full \ours{} & 48.8 \\
\bottomrule
\end{tabular}
}
\caption{Temporal-grounding performance under task removal.}
\label{tab:task_removal}
\end{table}

Table~\ref{tab:task_removal} shows that the remaining grounding tasks improve ActivityNet over
the SFT checkpoint, providing evidence of cross-task transfer. Direct temporal-grounding data
provides a further gain in the complete training mixture. The other representative metrics differ
from full \ours{} by at most 0.6.

\section{Expanded Component Ablations}

Table~\ref{tab:complete_ablation} expands the six representative benchmarks in
Table~\ref{tab:ablation_ema} to all of their evaluation metrics (e.g., R@0.3, R@0.5, R@0.7, and mIoU for ActivityNet). The ablation
configurations use the same model, data, training schedule, and rollout budget.

\begin{table}[h!]
\centering
\resizebox{0.48\textwidth}{!}{
\begin{tabular}{lcccc}
\toprule
& \multicolumn{2}{c}{\textbf{No declaration}} &
\multicolumn{2}{c}{\textbf{Generated declaration}} \\
\cmidrule(lr){2-3} \cmidrule(lr){4-5}
\textbf{Dataset and metric} &
\textbf{Per-task}$^\dagger$ & \textbf{Shared} &
\textbf{Per-task} & \textbf{Shared (\ours{})} \\
\midrule
\multicolumn{5}{l}{\textit{Spatial grounding: RefCOCO+ (Acc@0.5)}} \\
RefCOCO+ testA & 91.4 & 91.5 & 90.8 & 92.0 \\
RefCOCO+ testB & 82.7 & 82.6 & 82.1 & 82.9 \\
\midrule
\multicolumn{5}{l}{\textit{Temporal grounding: ActivityNet-Captions}} \\
ActivityNet R@0.3       & 65.0 & 66.7 & 68.3 & 68.7 \\
ActivityNet R@0.5       & 43.6 & 45.5 & 45.9 & 48.8 \\
ActivityNet R@0.7       & 25.7 & 27.1 & 27.4 & 28.1 \\
ActivityNet mIoU        & 45.9 & 47.0 & 47.8 & 48.2 \\
\midrule
\multicolumn{5}{l}{\textit{Spatio-temporal grounding (STVG)}} \\
tIoU@.5  & 34.9 & 35.1 & 35.8 & 35.6 \\
tIoU     & 39.5 & 39.2 & 39.4 & 39.8 \\
sIoU@.5  & 40.3 & 36.9 & 35.7 & 38.1 \\
sIoU     & 36.7 & 34.9 & 34.1 & 36.4 \\
\midrule
\multicolumn{5}{l}{\textit{Single-object tracking (GOT-10k)}} \\
AO    & 73.0 & 75.8 & 71.8 & 77.4 \\
R@0.3 & 93.9 & 92.8 & 95.0 & 96.1 \\
R@0.5 & 84.4 & 85.0 & 90.0 & 90.6 \\
R@0.7 & 68.8 & 66.1 & 75.0 & 75.6 \\
\midrule
\multicolumn{5}{l}{\textit{Image segmentation: RefCOCOg (cIoU)}} \\
RefCOCOg cIoU & 70.8 & 71.6 & 71.5 & 74.1 \\
\midrule
\multicolumn{5}{l}{\textit{Video segmentation: ReasonVOS}} \\
ReasonVOS J      & 51.1 & 54.1 & 53.4 & 55.8 \\
ReasonVOS F      & 58.7 & 61.6 & 61.4 & 63.7 \\
ReasonVOS J\&F   & 54.9 & 57.9 & 57.4 & 59.7 \\
\bottomrule
\end{tabular}}
\caption{Expanded metrics for the component ablations in Table~\ref{tab:ablation_ema}. $^\dagger$Values reported in OneThinker~\citep{feng2025onethinker}.}
\label{tab:complete_ablation}
\end{table}

\section{Validation on a Different Model Size}

Apart from the 8B model, we also validate \ours{} on the 4B model using the same training pipeline. As shown in Table~\ref{tab:progression_4b}, \ours{} consistently improves over the 4B base model across all task families, consistent with the 8B results.

\begin{table}[h!]
\centering
\resizebox{0.45\textwidth}{!}{
\begin{tabular}{l ccc}
\toprule
\textbf{Benchmark} & \textbf{Qwen3-VL-4B} & \textbf{+ SFT} & \textbf{+ RL} \\
\midrule
\multicolumn{4}{l}{\textit{Spatial Grounding (Acc@0.5)}} \\
RefCOCO testA & 90.6 & 92.0~\up{1.4} & 93.7~\up{1.7} \\
RefCOCO+ testA & 88.2 & 88.3~\up{0.1} & 91.6~\up{3.3} \\
RefCOCOg test & 85.1 & 88.6~\up{3.5} & 89.0~\up{0.4} \\
\midrule
\multicolumn{4}{l}{\textit{Temporal Grounding}} \\
Charades R@0.5 & 35.9 & 65.7~\up{29.8} & 71.3~\up{5.6} \\
\midrule
\multicolumn{4}{l}{\textit{Tracking}} \\
GOT-10k R@0.5 & 41.7 & 46.7~\up{5.0} & 89.4~\up{42.7} \\
\midrule
\multicolumn{4}{l}{\textit{Spatio-Temporal Grounding}} \\
STVG tIoU@0.5 & 21.3 & 33.5~\up{12.2} & 34.3~\up{0.8} \\
\midrule
\multicolumn{4}{l}{\textit{Image Segmentation (cIoU)}} \\
RefCOCO & 74.8 & 75.6~\up{0.8} & 75.4~\dn{0.2} \\
RefCOCO+ & 66.9 & 65.1~\dn{1.8} & 68.0~\up{2.9} \\
RefCOCOg & 69.6 & 70.7~\up{1.1} & 71.9~\up{1.2} \\
\midrule
\multicolumn{4}{l}{\textit{Video Segmentation (J\&F)}} \\
MeViS & 22.7 & 53.3~\up{30.6} & 56.0~\up{2.7} \\
ReasonVOS & 24.5 & 54.3~\up{29.8} & 57.5~\up{3.2} \\
\midrule
\multicolumn{4}{l}{\textit{Video QA}} \\
VideoMMMU & 58.6 & 59.3~\up{0.7} & 62.4~\up{3.1} \\
\bottomrule
\end{tabular}}
\caption{Validation of our approach on 4B model: base model $\to$ SFT with task-aware data $\to$ RL with unified normalization and task-ID reward. Small colored numbers are the change of each stage over the previous one, as in Table~\ref{tab:progression}.}
\label{tab:progression_4b}
\end{table}

\section{Zero-Shot Multi-Context Grounding}

We evaluate the existing checkpoint on the 1,708-sample MC-Bench validation
set~\citep{mcbench} without using multi-image grounding data during training.
\begin{table}[h!]
\centering
\resizebox{0.5\textwidth}{!}{
\begin{tabular}{lrrrrrrrr}
\toprule
& \multicolumn{4}{c}{\textbf{Image-level Acc@0.5}} &
\multicolumn{4}{c}{\textbf{Instance-level AP50}} \\
\cmidrule(lr){2-5} \cmidrule(lr){6-9}
\textbf{Model} & \textbf{Ref.} & \textbf{Com.} & \textbf{Rea.} & \textbf{All} &
\textbf{Ref.} & \textbf{Com.} & \textbf{Rea.} & \textbf{All} \\
\midrule
GPT & \best{69.4} & \best{82.8} & \best{77.5} & \best{78.3} &
1.8 & 4.1 & 2.3 & 2.8 \\
Gemini & 55.8 & 65.1 & 62.7 & 62.5 &
30.5 & 30.0 & 26.1 & 28.4 \\
\ours{} & 62.6 & 67.9 & 61.8 & 64.5 &
\best{55.0} & \best{60.2} & \best{54.0} & \best{56.7} \\
\bottomrule
\end{tabular}}
\caption{Zero-shot evaluation on the MC-Bench validation set. Ref., Com., and Rea. denote
referring, comparison, and reasoning prompts, respectively.}
\label{tab:mcbench}
\end{table}

\ours{} exceeds Gemini but remains below GPT on overall image-level accuracy. At the
instance level, \ours{} obtains the highest AP50 across all three prompt styles. These results
support zero-shot multi-image grounding compatibility rather than trained specialization.

\section{Failure Cases Analysis}

\begin{figure}[h!]
\centering
\begin{minipage}[t]{0.48\textwidth}
\centering
\includegraphics[width=\linewidth]{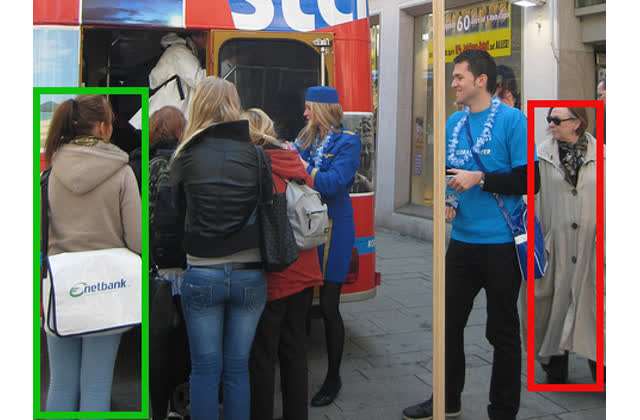}\\[-2pt]
\small\textbf{(a)} ``brown coat.''
\end{minipage}
\hfill
\begin{minipage}[t]{0.48\textwidth}
\centering
\includegraphics[width=\linewidth]{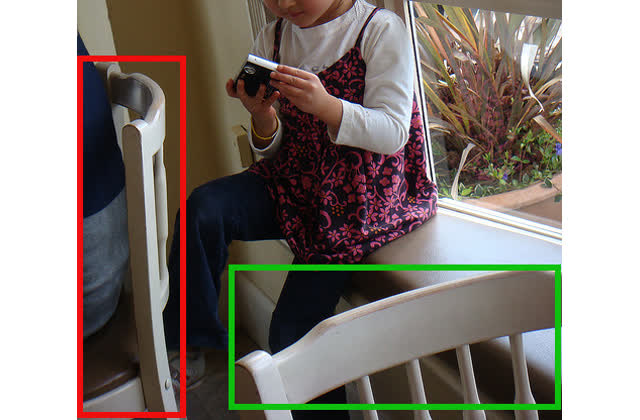}\\[-2pt]
\small\textbf{(b)} ``chair partly cut off by corner of picture.''
\end{minipage}

\vspace{6pt}
\begin{minipage}[t]{0.48\textwidth}
\centering
\includegraphics[width=\linewidth]{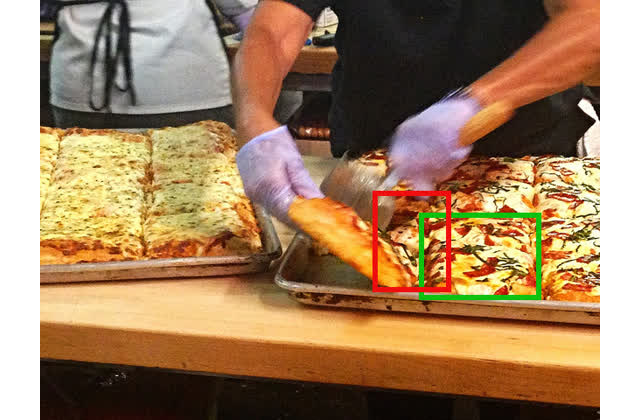}\\[-2pt]
\small\textbf{(c)} ``slice next to uplifted piece.''
\end{minipage}
\hfill
\begin{minipage}[t]{0.48\textwidth}
\centering
\includegraphics[width=\linewidth]{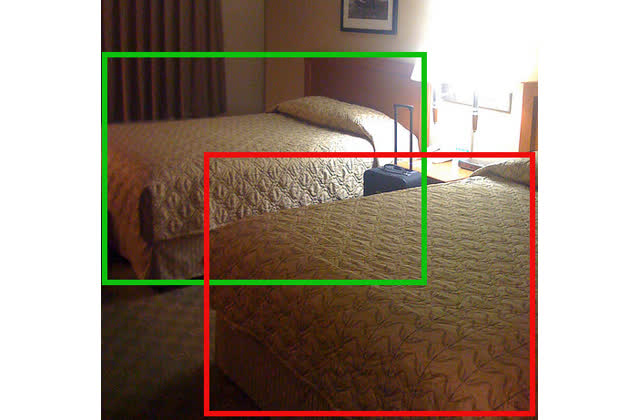}\\[-2pt]
\small\textbf{(d)} ``bed with light shine on.''
\end{minipage}
\caption{Representative spatial-grounding failures. Green and red denote the ground-truth and
predicted bounding boxes, respectively.}
\label{fig:failure_cases}
\end{figure}

Figure~\ref{fig:failure_cases} presents four representative spatial-grounding failures. The model selects an incorrect instance or
misinterprets the referring relation.

In (a), the model selects another person with a similarly colored coat. In (b), it selects the
chair on the left rather than the partially cropped foreground chair. In (c), it selects the
uplifted piece itself rather than the adjacent slice. In (d), it selects the foreground bed
rather than the illuminated rear bed. These cases expose remaining challenges in fine-grained
instance disambiguation and relational grounding.

\end{document}